\documentclass[runningheads]{llncs}
\usepackage[T1]{fontenc}
\usepackage{cite}
\usepackage{graphicx}
\usepackage{hyperref}
\usepackage{multirow}
\usepackage{placeins}
\usepackage{longtable}
\usepackage{threeparttable}
\usepackage{amsmath}
\usepackage{array}
\usepackage{booktabs}
\usepackage[table]{xcolor} 
\usepackage{tabularx}
\graphicspath{{figs/}}
\usepackage{color}

\begin{document}
\title{TractoEmbed: Modular Multi-level Embedding framework for white matter tract segmentation \thanks{Supported by SERB (Science and Engineering Research Board) of India}}

\titlerunning{TractoEmbed}
\author{Anoushkrit Goel \inst{1} \and
Bipanjit Singh \inst{1} \and
Ankita Joshi\inst{1} \and
Ranjeet Ranjan Jha\inst{2} \and
Chirag Ahuja\inst{3} \and
Aditya Nigam\inst{1} \and
Arnav Bhavsar\inst{1}
}
\titlerunning{TractoEmbed}
\authorrunning{A. Goel et al.}

\institute{Indian Institute of Technology (IIT) Mandi, India \and
Indian Institute of Technology (IIT) Patna, India \and
Post-Graduate Inst. of Medical Edu. and Research (PGIMER), Chandigarh, India}
\maketitle
\begin{abstract}
White matter tract segmentation is crucial for studying brain structural connectivity and neurosurgical planning. 
However, segmentation remains challenging due to issues like class imbalance between major and minor tracts, structural similarity, subject variability, symmetric streamlines between hemispheres etc.
To address these challenges, we propose TractoEmbed, a modular multi-level embedding framework, that encodes localized representations through learning tasks in respective encoders.
In this paper, TractoEmbed introduces a novel hierarchical streamline data representation that captures maximum spatial information at each level i.e. individual streamlines, clusters, and patches.
Experiments show that TractoEmbed outperforms state-of-the-art methods in white matter tract segmentation across different datasets, and spanning various age groups. The modular framework directly allows the integration of additional embeddings in future works. 
\end{abstract}
\section{Introduction} \label{intro}
\textbf{Diffusion MRI (dMRI)} \cite{basser2000vivo,basser1994mr} facilitates the non-invasive examination of the brain's white matter (WM) microstructural organization. A crucial component of the dMRI analysis pipeline is fiber tractography \cite{behrens2007probabilistic,tournier2004direct,tournier2007robust}, which tracks fibers or streamlines under anatomical constraints from the dMRI signal received from the scanner(refer to Section \ref{data:diffusion}).
Tract Segmentation involves dividing the streamlines into distinct, anatomically meaningful tracts, with each tract corresponding to a specific white matter pathway. These tracts can be broadly grouped into 3 types based on structural connectivity, i.e. Association, Commissural, and Projection Fibers. Each type is further subdivided based on its specific structural connectivity and function, allowing for more granular distinctions.
Through the segmentation process, it becomes possible to conduct quantitative studies of white matter (WM), which is important in understanding neurological disorders such as Alzheimer's, and Parkinson's \cite{marek2011parkinson}, the effect of tumors on segmenting fiber streamlines, etc.

In addition, tract segmentation is also crucial for preoperative neurosurgical planning \cite{lucena2022informative}, as it helps identify eloquent white matter areas and determine optimal surgical approaches that minimize post-operative damage. Tract segmentation is also extensively used to visualize particular segments for focused examination by clinicians. However, this process is typically performed by expert Neuroanatomists using their knowledge of brain anatomy to divide fibers into multiple bundles. As a result, it is very time-consuming and can vary between experts, affecting the consistency and reliability of the results.

Taking challenges associated with manual tract segmentation, various techniques have been developed over the years.  These techniques range from classical methods to ATLAS-based and distance-based algorithms \cite{garyfallidis2018recognition,garyfallidis2012quickbundles,st2022fast,vindas2023geolab} (refer to Section \ref{related-work}).
An ATLAS refers to a standardized reference that allows spatial mapping of neuroimaging data from different studies (refer to Table \ref{tab:datasets}) and modalities. They approximate the shape, location, and brain region boundaries in a common coordinate space, facilitating the comparison of brain structure and function across individuals.

These methods require significant manual intervention and are prone to age-related brain changes, also their effectiveness depends on the alignment and quality of the ATLAS.
Considering the limitations of manual and classical methods, as well as the importance of tract segmentation, machine learning, and deep learning-based frameworks have been proposed for automatic tract segmentation \cite{zhang2020deep, berto2021classifyber, wasserthal2018tractseg}. Deep learning algorithms can learn information from shape, structure, relative location, fiber orientations, etc. 

However, a notable drawback is that these models often fail in classifying streamlines that are linear in shape due to over-reliance on shape, such as striato-thalamo-pallido projection fibers, which in existing methods, require global reference along with streamlines \cite{funk2023humans}. 
Additionally, when neurosurgeons are concerned with segmenting only a specific set of streamlines, global tractography can become a computational overhead.
Due to these complexities in streamline classification-based tract segmentation, each method inherently has a certain drawback. 

To address this, we propose TractoEmbed, a modular framework that combines multi-level embeddings extracted from hierarchical data representations specifically at streamline, patch, and cluster levels (refer to Fig. \ref{fig:data-representation}). Our approach surpasses state-of-the-art (SOTA) results in tract segmentation.
In this work, we present an approach with the following major contributions:
\begin{enumerate}
\item We introduce \textbf{TractoEmbed}, a \textbf{novel modular multi-embedding framework}, which leverages learning task-specific encoders to embed data representations, and generate embeddings. Where the encoders and their hyperparameters are selected after rigorous experimentation.
\item We propose \textbf{novel hierarchical and descriptive streamline data representations}. These representations includes spatial information about regional patches, neighboring streamlines and the streamline itself, providing a comprehensive understanding of the streamline characteristics. 
In contrast to recent advances, our method leverages minimal neighbouring streamlines, hyperlocal streamlines, enhancing robustness to practical clinical settings
\item It is demonstrated that TractoEmbed framework, \textbf{generalizes across various datasets} encompassing different age groups (refer Table \ref{tab:datasets}). 
Additionally, the framework is \textbf{modular} at the embedding level, allowing researchers to \textbf{integrate their own learnable embeddings} to achieve even richer representations of streamline data. 
This modularity enhances the flexibility and adaptability of the framework.
\end{enumerate} 
\section{Related Work} \label{related-work}
In recent years, a plethora of classical and deep learning methods have been developed for tract segmentation, capable of performing in diverse conditions and data formats with minimal supervision from the skilled medical practitioners.
Among these methods, clustering and distance-based methods, QuickBundles \cite{garyfallidis2012quickbundles} and RecoBundles \cite{garyfallidis2018recognition} are fast algorithms that utilize clustering approaches. QuickBundles is known for its speed in grouping streamlines based on their similarity, while RecoBundles excels in identifying parent anatomical bundles of streamlines. RecoBundles achieves this by recognizing and clustering similar streamlines based on their shape and spatial location, meanwhile leveraging a model of known white matter anatomy for accurate segmentation. 
Additionally, the Fast Streamline Search (FSS) \cite{st2022fast} is a highly accurate distance-based search method. FSS indexes streamlines in a spatial data structure, enabling efficient retrieval of similar streamlines in tractography data.

Other notable methods include GeoLab \cite{vindas2023geolab}, a tract segmentation framework for analyzing the geometry, topology, and structural connectivity of white matter fiber bundles. Classifyber \cite{berto2021classifyber} is a linear classifier that uses distance-based embeddings with local and global streamlines and regions of interest (ROIs) in the brain, concatenated into a weight vector, which serves as a hybrid of distance-based and learning-based algorithms.
TractSeg \cite{wasserthal2018tractseg}, one of the seminal works, uses a 2D U-Net model that directly works on fODF peaks \cite{tournier2004direct} to segment tracts, without the need for parcellation and registration.
In DeepWMA \cite{zhang2020deep}, shape information of a single streamline is used to feed a FiberMap to a simple CNN model, preserving local information. BrainSegNet \cite{gupta2017brainsegnet} employs bi-directional LSTMs, while FS2Net \cite{jha2019fs2net} uses an LSTM-based model to develop a rotation-invariant segmentation model. 
TRAFIC \cite{lam2018trafic} uses geometry and 265 landmarks to accurately label, classify, and clean the traced paths of streamlines in streamline space. 

Xue \textit{et al.} use the PointNet model to classify streamlines using a local-global data representation, and Wang \textit{et al.} \cite{wang2022accurate} utilize a transformer encoder for fiber segmentation by incorporating features related to fiber shape and position. 
In \cite{liu2019deepbundle}, a graph convolution (GCNN)-based framework, Spectral GCNN extracts geometry-invariant features. 
In FIESTA, \cite{dumais2023fiesta} Dumais \textit{et al.} segment tracts in latent space, via an autoencoder-based segmentation algorithm.
\section{Diffusion MRI Data}\label{data:diffusion}
In this section, we discuss how diffusion MRI data (refer to Table \ref{tab:datasets}) is acquired and processed to generate input and labels for the proposed framework. Additionally, we explain how the data is divided for training and testing purposes and converted to variations of Point Cloud Data before feeding to encoders.
\begin{table}[!hbt]
\caption{Description of the publicly available dMRI Datasets containing a total of 1 million streamlines with (15,3) dimension each i.e. (1000000, 15, 3) using UKF \\Tractography and Parcellation (refer Section \ref{subsec:tractography}. \cite{xue2023tractcloud}} 
\label{tab:datasets}
\centering
\begin{tabular}{ p{4cm}|p{0.75cm}|p{1cm}|p{1.4cm}|p{1.5cm}|c }
 \hline
 \textbf{Neuroimaging Datasets}& \textbf{N subs} &\textbf{b $s/mm^2$} &\textbf{N Volumes} ($mm^3$)&\textbf{TE/TR} ($ms$)&\textbf{Resolution} ($mm^3$)\\
 \hline
 \textbf{dHCP \cite{edwards2022developing}}   && 0& 20 vol.& 90/3800  & 1.5x1.5x1.5 mm3  \\
\cline{3-4}
 developing Human& 20 & 400& 64 vol. &&\\
\cline{3-4}
Connectome Project & &1000& 88 vol. &&\\
\cline{3-4}
 &&2600& 128 vol. &&\\
 \hline
 \textbf{ABCD \cite{volkow2018conception}} &25 & 0 & 1 vol. & 88/4100 & 1.7x1.7x1.7  \\
\cline{3-4}
 Adolescent Brain Cognitive Development & &3000& 60 vol.  &&\\
 \hline
 \textbf{HCP \cite{van2013wu}} & 25 & 0 & 18 vol.& 89/5520 &1.25x1.25x1.25 \\
\cline{3-4}
 Human Connectome Project &  & 3000& 90 vol. &&\\
 \hline 
\textbf{PPMI \cite{marek2011parkinson}} Parkinson's& 25 &0 & 1 volume & 88/7600 & 2x2x2    \\
\cline{3-4}
Progression Markers Initiative &&1000& 64 vol.  &&\\
 \hline
\textbf{BTP \cite{zhang2020deep}} Brigham's Tumor &25&0 & 1 volume   & 98/12700 &  2.2x2.2x2.3 \\
\cline{3-4}
Patient Data&&2000& 30 vol. &&\\
 \hline
\end{tabular}
\label{tab:datasets}
\end{table}

\subsection{Data Preparation} \label{subsec:tractography}
Diffusion MRI data \cite{basser2000vivo} (refer to Table \ref{tab:datasets}) is acquired by applying magnetic diffusion gradients and measuring the resulting signal attenuation, which depends on the local tissue microstructure. This diffusion MRI is preprocessed using standardized algorithms \cite{tournier2007robust,tournier2004direct}, followed by streamlines tracking using a tractography algorithm \cite{malcolm2009neural,behrens2007probabilistic}. ATLAS based labelling of streamlines is performed in the parcellation process. 
ATLAS registration on different brains can be inconsistent, non-scalable, knowledge intensive, dataset-specific and time-consuming because they are created by expert neuroanatomists.
Hence there is a need for algorithms to automate tract segmentation.

For \textbf{Tractography}, we utilize the Unscented Kalman Filter (UKF) \cite{malcolm2009neural}, which estimates microstructural parameters and fiber orientations from diffusion MRI data to track neuronal paths from multiple seed points. After tractography, the extracted streamlines are bundled into parcels or clusters, where these parcels are mapped to anatomically meaningful tracts using ATLAS, as discussed below. 

\label{subsec:parcellation} \textbf{Parcellation} refers to the division of the brain into anatomical regions based on ATLAS and clustering techniques into parcels. 
Initially, division of hemisphere in streamline space, registration on ATLAS, and transformations are performed to align current brain with the ATLAS. 

Through this process, we obtain 800 parcels that are appended to anatomical tracts and labeled along with quality control. 
These parcels are further refined using the ATLAS and diffusion measurements to separate outlier streamlines from each parcel, dividing each parcel into 2, resulting in 800 outlier parcels and 800 plausible parcels. 

Using ATLAS, we club and label all 800 Outlier parcels to "Other" label, and 800 plausible parcels to 42 anatomical tracts. This entire procedure can be executed using \textit{whitematteranalysis} package \cite{zhang2018anatomically,o2007automatic,o2012unbiased}, which follows these steps from tractography streamlines to parcellation. This method ensures consistency across subjects and datasets.
ATLAS used in this paper was derived from mean of \textit{100 registered tractography of young healthy adults in the Human Connectome Project (HCP)} \cite{van2013wu}.

\subsection{Training and Testing Data } \label{train-test}
After sequentially performing fiber tractography, parcellation, and labeling of each parcel, we obtain a total of 1 million labeled streamlines from 100 out of 120 subjects (refer to Table \ref{tab:datasets}), where each subject contains 10,000 streamlines. And 20 subjects out of 120 subjects are kept aside for real world test cases, and not included in data splits.

From a total corpus of 100 subjects, we obtain an array consisting of 1M streamlines of shape \textbf{(1000000, 15, 3)} (refer Table \ref{tab:datasets}), where (15,3) streamline array is derived from feature data in RAS (Right, Anterior, Superior) coordinate space (refer Supplementary Material). 
This dataset is subsequently partitioned into \textbf{train, validation, and test} sets in a ratio of 70 subjects for training, 10 subjects for validation, and 20 subjects for testing. Data is split subject-wise, where a subject will only belong to one data split at a time\cite{xue2023tractcloud}.

The dataset encompasses 43 tract classes: 42 anatomically significant tracts spanning the entire brain, and one category labeled as “other", which includes anatomically implausible outlier streamlines identified during the parcellation process (refer Section \ref{subsec:parcellation}). Here \textbf{PCD} is an acronym for Point Cloud Data.

\begin{figure}[!hbt] 
    \centering
    \includegraphics[width=1\linewidth]{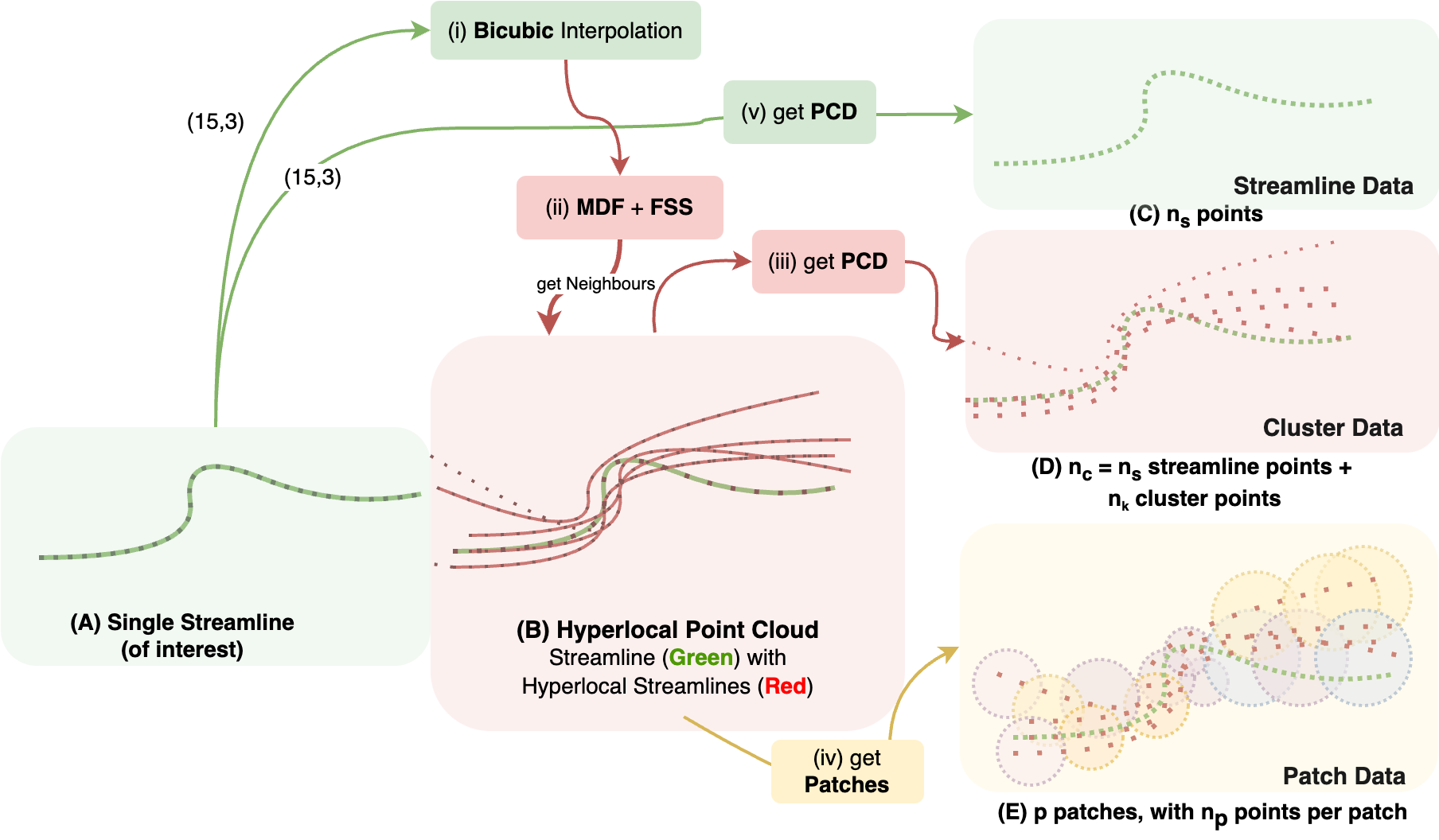}
    \caption{\textbf{Data Representations: Streamline, Patch, and Cluster}. 
    For \textbf{(C)} Streamline Data, \textbf{(A)}input streamline of shape (15,3), is (v) converted to a point cloud.
    For \textbf{(B)} Hyperlocal Streamlines, (refer Section \ref{sec:pcd}) input streamline undergoes (i) bicubic interpolation to make a streamline of shape (40,3), on which $k_{local}$ neighboring streamlines are sampled using (ii) MDF Distance. In $k_{local}$ search space, FSS \cite{st2022fast}, (ii) Fast Streamline Search is used to get 5 \textbf{(B)} hyperlocal streamlines.
    (iii) \textit{get PCD} converts hyperlocal streamlines to\textbf{ (D)} Cluster Data with ($n_c$,3) points.
    For\textbf{ (E)} Patch Data, (iv) $p$ farthest points are sampled using FPS (refer Section \ref{subsec:patch-point-cloud}), $n_p$ points in each patch using kNN to find neighboring points.}
    \label{fig:data-representation}
\end{figure}
\subsection{Model Input Data} \label{sec:pcd}
Training and Testing Data (refer Section \ref{train-test}), is in the form of a three-dimensional array that contains (number of streamlines, points per streamline, number of features) and is in unusable form for most encoders. Hence, to make it suitable for encoder-specific preprocessing methods, we represent data in 3 forms (as mentioned in Fig \ref{fig:data-representation}). which would be utilised by encoders in sections \ref{sembed}\ref{cembed}\ref{pembed} and in results section \ref{sec:results}.

\begin{enumerate}
\item \textbf{Streamline Data}: As mentioned in Fig \ref{fig:data-representation}, Streamline Data is a (15, 3) array, which is created by undersampling actual streamlines of different lengths, containing RAS coordinates as features (also refer Section \ref{sembed}). 

\item \textbf{Local Point Cloud} (\textit{Local PCD}): Streamlines are interpolated using bicubic interpolation to approximate a smooth curve of the streamline, experimentally tested to be a (40,3) streamline. 
On the interpolated streamline, we use \textbf{MDF} (Mean Direct Flip) \cite{garyfallidis2012quickbundles} distance to find $k_{local}$ neighboring local streamlines. These $k_{local}$ streamlines are then converted to \textbf{Local Point Cloud} (Local PCD) by merging and randomly sampling $n_c$ \textit{(number of points in a cluster)} points from $((k_{local}+1)*40, 3)$ Here interpolated streamlines give dense point clouds, giving richer representation.  

\item \textbf{Hyperlocal Point Cloud} (\textit{Hyperlocal PCD}): In the limited search space of $k_{local}$ Local Streamlines, we employ \textbf{FSS} (Fast Streamline Search)\cite{st2022fast} with a radius of 4mm-6mm to find 5 closest streamlines to the streamline of interest. This group of 5 hyperlocal streamlines is then converted to make a Hyperlocal Point Cloud by merging and randomly sampling $n_c$ points from total points of dimensions (240,3), from $((5 + 1) * 40, 3)$ which is (240,3), where 5 is no. of hyperlocal streamlines and 1 is the streamline itself.
\end{enumerate}

Hyperlocal Point Cloud is a variation of Local Point Cloud which contains fewer spatially similar streamlines wherein local streamlines in Local PCD can range to higher numbers also, and may contain dissimilar streamlines with neighboring spatial information rather than structural shape information. 
Models we have used as respective encoders can utilize certain forms of data. \textbf{Streamline Encoder} can only process Streamline Data. \textbf{Patch Encoder} can process Hyperlocal and Local PCD. \textbf{Cluster Encoder} can process all forms of data mentioned above (refer Fig \ref{fig:data-representation}).

\section{Methodology} \label{sec:method}
\textbf{TractoEmbed} utilizes a modular framework to fuse learnable embeddings trained on hierarchical streamline data representations, as detailed in the following subsections: \ref{sembed}, \ref{cembed}, and \ref{pembed}. 

\begin{figure}[!ht]
    \centering
    \includegraphics[width=1\linewidth]{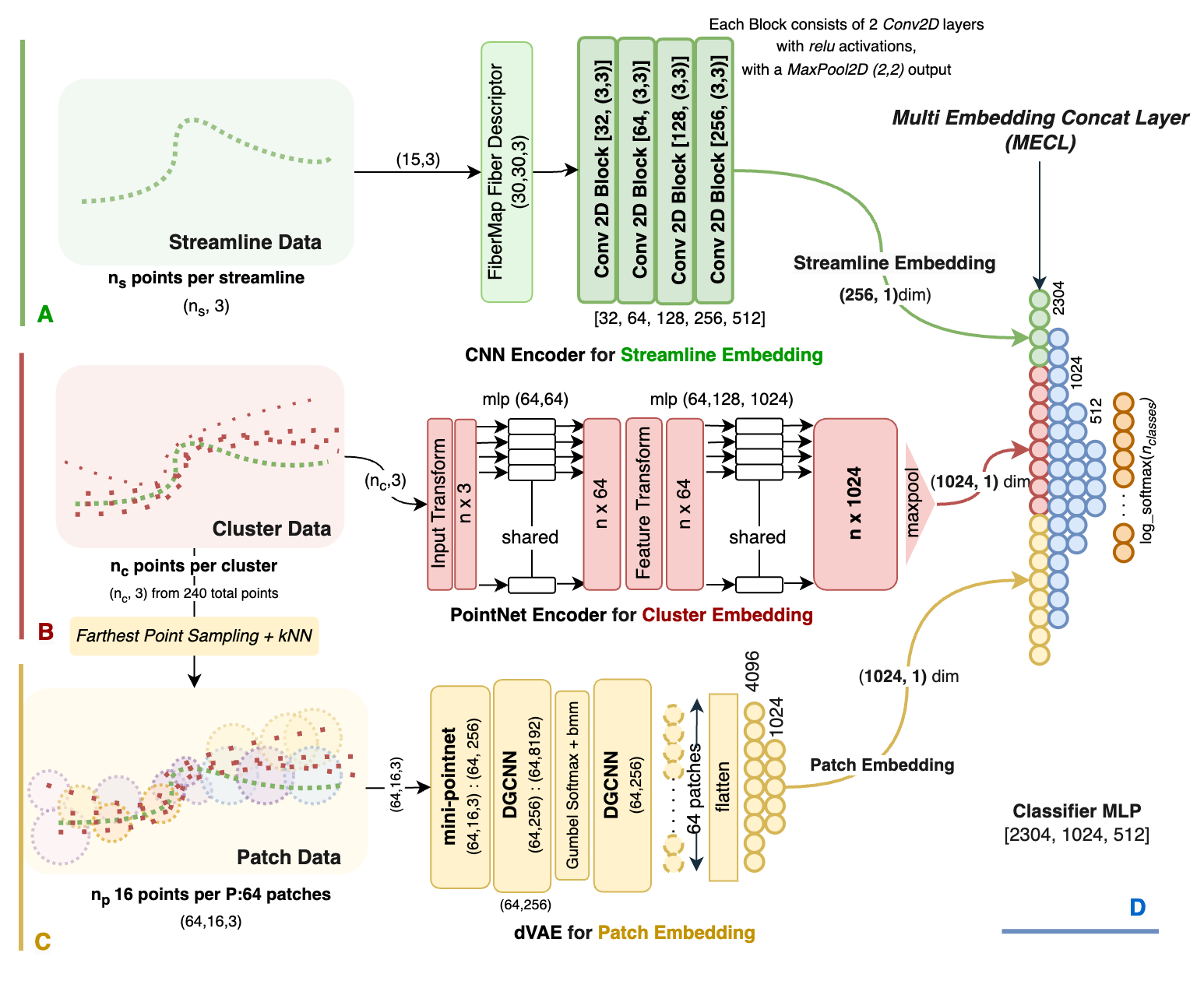}
    \caption{Streamline, Patch, and Cluster data obtained through the processes illustrated in Fig.: \ref{fig:data-representation}, are sent to respective encoders to generate embeddings.\\
    \textbf{(A)}\textit{\textbf{Streamline Data}} of dimensions  ($n_s$, 3) serves as input to the Fiber Descriptor \cite{zhang2020deep}, producing an output of dimensions ($2*n_s$, $2*n_s$, 3), where $n_s$ is number of points per streamline, which is fed to 4 CNN blocks, refer Table \ref{tab:sencoder}, to obtain a final embedding of dimensions (256, 1) for the \textit{MECL (Multi Embedding Concat Layer)}. \\
    \textbf{(B)} \textit{\textbf{Cluster Data}} $(n_c,3)$ from either local or hyperlocal point cloud, refer to section \ref{sec:pcd}, is fed to the PointNet Encoder to give cluster embedding of 1024 dimensions.\\ 
    Patches on Cluster Data are created using Farthest Point Sampling to fetch 64 patches with 16 points each, resulting in (64,16,3) dimensions. \\
    \textbf{(C)} \textit{\textbf{Patch data}} is fed to a mini PointNet, which produces a (64,256) output, further input to dVAE, resulting in 64 patches each of dimension 256. This is flattened to be fed to [4096, 1024] dense layers to give an output patch embedding of 1024 dimensions.\\
    \textbf{(D)} These multiple embeddings are concatenated at \textit{MECL} to make (256 + 1024 + 1024 = 2304) dimensional embedding. This is input to Classifier MLP resulting in a 512 dim classification embedding.}
    \label{fig:multi-embedding-model}
\end{figure}
In the Methodology section we describe pre-processing, training method, model architecture, and output embedding for each encoder.
\begin{enumerate}
    \item \textbf{Streamline Encoder} essentially is any model that preserves intra-streamline information, its order of points, shape, and geometry amidst the random shuffling of data points in other models.
    To preserve intra-streamline spatial information we chose CNN-based method due CNN's inherent capability of learning local and global features from a 2D array with channels. One can argue LSTM encoder for auto-regressive sequential information but LSTM struggles to encode spatial information (refer Section \ref{sec:results})
    \item \textbf{Cluster Encoder}, should encode the shape, inter-streamline dependencies, and information of a cluster to resemble the target tract. Based on our evaluation PointNet is imperative in understanding spatial features from a cluster of points or point cloud. We did mild variations in kernel sizes and layers. We found the simple PointNet\cite{qi2017pointnet} model's ability to discern intricate patterns and dependencies better than others.
    \item Objective of \textbf{Patch Encoder} is to learn regional information in a hyperlocal streamline point cloud, to embed origin and termination region information in the point cloud through regional patches. We chose a combination of mini-pointnet and Discrete Variational Autoencoder (dVAE) \cite{rolfe2016discrete} to reconstruct point cloud patches and learn regional generative features. 
    Patches are used to embed regional information as attention across only points fails due to minimal information in a single point and high compute requirements\cite{yu2022point}.

\end{enumerate}

Broadly, three types of encoders are pre-trained or finetuned for classification downstream tasks. Embeddings from these encoders are combined to assist the classifier MLP (as illustrated in Fig. \ref{fig:multi-embedding-model}) in achieving accurate classification.
\subsection{Streamline Encoder} \label{sembed}

\textbf{Preprocessing:}
The streamline(15,3) is passed through the Fiber Descriptor to get streamline representation for streamline encoder. Fiber Descriptor is a streamline representation technique where concatenation of normal and flipped streamlines are stacked, so that a CNN kernel can learn local intra-streamline features. 

In Fiber descriptor, the input streamline is flipped and concatenated horizontally with the original (15,3) input, creating a (30,3) row. In the second row, the flipped streamline is followed by the original, forming a 12-21 pattern (where 1 represents the original and 2 represents the flipped). This 12-21 pattern (representation of 2 rows separated by a dash `-') is vertically repeated 15 times, resulting in (30,30,3) representation from (15,3). This is input to our streamline encoder model (for more details refer \cite{zhang2020deep}).

\textbf{Streamline Embedding} is the 256-dimensional output generated by the streamline encoder model (refer to Table \ref{tab:sencoder}). This CNN model is independently trained on the Fiber Descriptor representation of streamlines using cross-entropy loss for classification. The model is designed to learn streamline-specific discriminative spatial features, (as discussed \cite{zhang2020deep}). We extract the streamline embedding from the output of \textbf{MaxPool2D} layer (refer Fig. \ref{fig:multi-embedding-model} and Table \ref{tab:sencoder}).

\begin{table}[!ht]
    \centering
    \caption{\textbf{Streamline Encoder}: Stacked CNNs with input data size (30,30,3), which we have compressed to represent as 4 Conv Blocks }
    \label{tab:sencoder}
    \begin{tabular}{lllp{4cm}}
        \toprule
        \textbf{Layer (type)} & \textbf{Output Shape} & \textbf{Kernel} & \textbf{Configuration} \\
        \midrule
        \textbf{Conv Block 1} & (None, 30, 60, 32) & (3, 3) & [Conv2D, Activation] x2 \\
        Max Pooling 2D& (None, 15, 30, 32) & (2, 2) & - \\
        \textbf{Conv Block 2} & (None, 15, 30, 64) & (3, 3) & [Conv2D, Activation] x2 \\
        Max Pooling 2D & (None, 7, 15, 64) & (2, 2) & - \\
        Dropout & (None, 7, 15, 64) & - & - \\
        \textbf{Conv Block 3} & (None, 7, 15, 128) & (3, 3) & [Conv2D, Activation] x2 \\
        Max Pooling 2D & (None, 3, 7, 128) & (2, 2) & - \\
        \textbf{Conv Block 4} & (None, 3, 7, 256) & (3, 3) & [Conv2D, Activation] x2 \\
        Max Pooling 2D & (None, 1, 3, 256) & (2, 2) & \textit{Streamline Embedding} \\
        \bottomrule
    \end{tabular}
\end{table}
\subsection{Cluster Encoder} \label{cembed} \label{subsec:cluster-point-cloud} 
\textbf{Preprocessing:} From interpolated (40,3) streamlines we sample $k_{local}$ local neighbour streamlines using MDF Distance from QuickBundles. After finding local streamlines for each streamline, we sample hyperlocal streamlines using \textbf{FSS} \cite{st2022fast}. FSS uses barycenter of streamlines and distance parameters like radius to accurately find similar streamlines. Streamlines sampled from FSS are highly probable to belong to the same class which allow us to merge these streamlines creating hyperlocal streamline data (6,40,3) where $k_{hyperlocal}$=5 and 1 streamline, resulting in (240,3) point cloud, from which $n_c$ points are sampled for the input to PointNet Model (refer Table \ref{tab:cencoder}) for the Cluster Embedding.
\begin{table}[!htp]
\centering
\caption{\textbf{Cluster Encoder}: PointNet Model Architecture for Cluster Embedding, extracted at the last layer in this table.}
\label{tab:cencoder}
\begin{threeparttable}
\begin{tabular}{lrrrrr}
\toprule
\textbf{Model} & \textbf{Module} & \textbf{$C_{in}$} & \textbf{$C_{out}$} & \textbf{Kernel Size} \\
\cmidrule{1-5}
\multirow{5}{*}{\textbf{PointNet Encoder}} & Conv1D and BN & 3 & 64 & 1 \\
\cmidrule{2-5}
& STNkD & 64 & 64 & - \\
\cmidrule{2-5}
& Conv1D and BN & 64 & 128 & 1 \\
\cmidrule{2-5}
& Conv1D and BN & 128 & 256 & 1 \\
\cmidrule{2-5}
& Conv1D and BN & 256 & 1024 & 1 \\
\midrule
\end{tabular}
\end{threeparttable}
\end{table}

\textbf{Cluster Embedding}\label{cembed} is a 1024 dimensional output embedding of a PointNet Model (refer Table \ref{tab:cencoder}\cite{qi2017pointnet}), which takes in Cluster data (refer \ref{subsec:cluster-point-cloud}) as input ($n_c$,3), where $n_c$ is the number of points in the total point cloud.

Cluster Encoder architecture is mentioned in Table \ref{tab:cencoder}, all values are Xavier initialized, and the Cluster Embedding (1024 dimensional) is extracted at final MaxPool 1D. Cluster Encoder is trained along with Classifer MLP of TractoEmbed (discussed in  \ref{subsec:training-tractoembed}).Effects on Classification Accuracy are observed by varying, $n_c$ in Ablation Study Table \ref{tab:ablation-table} 

\subsection{Patch Encoder} \label{pembed}
\textbf{Patch Data} \label{subsec:patch-point-cloud} refers to a 3D patch on the Cluster data. We use patch data to capture information across and among regional patches on a group of streamlines (Cluster). Enabling us to classify streamlines that are structurally linear in shape, and are very similar to other streamlines (like projection fibers). Regional Patches will learn to focus on Origin and Termination ROIs to help better segment difficult tracts.

\textbf{Preprocessing:} Patches are created by iteratively sampling, $p_f$ farthest points using FPS, Farthest Point Sampling, over the point cloud. we then use kNN to sample $p_{local}$ nearest points per patch. We get $p_f=64$ patches, where every patch has $p_{local}=16$ points making patch data dimensions to be (64, 16, 3) from an input cluster data of $(n_c,3)$, randomly sampled with replacement. (refer Ablation Study \ref{tab:ablation-table} to see effects on variation in $n_c$)
\begin{table}[!hbt]
\centering

\caption{\textbf{Patch Encoder}: dVAE model architecture details \cite{yu2022point}. The last layer from which Patch Embedding is extracted. Where 
$C_{in}$ represents dimensions of input features, and $C_{out}$, dimensions of output features. $N_{out}$ is the number of points in the query point cloud. $K$ is the number of neighbors in \textbf{kNN} operation. $C_{middle}$ is the dimension of the hidden layers for MLPs.}
\label{tab:patch-encoder}
\begin{threeparttable}
\begin{tabular}{p{2cm}|p{1.5cm}|p{2cm}|p{1cm}|p{1cm}|p{1cm}|p{1cm}|c}
\hline 
\textbf{Model} & \textbf{Module} & \textbf{Block} & \textbf{$C_{in}$} & \textbf{$C_{out}$} & \textbf{$K$} & \textbf{$N_{out}$} & \textbf{$C_{middle}$} \\
\hline 
\multirow{12}{*}{\textbf{dVAE}} 
 & \multirow{6}{*}{Encoder} 
  & Linear & 256 & 128 & - & - & - \\
 &  & DGCNN & 128 & 256 & 4 & 64 & - \\
 &  & DGCNN & 256 & 512 & 4 & 64 & - \\
 &  & DGCNN & 512 & 512 & 4 & 64 & - \\
 &  & DGCNN & 512 & 1024 & 4 & 64 & - \\
 &  & Linear & 2304 & 8192 & - & - & - \\
\cline{2-8}
 & \multirow{8}{*}{Decoder} 
  & Linear & 256 & 128 & - & - & \multirow{8}{*}{\begin{tabular}{c} 1024 \\ 1024 \end{tabular}} \\
 &  & DGCNN & 128 & 256 & 4 & 64 & \\
 &  & DGCNN & 256 & 512 & 4 & 64 & \\
 &  & DGCNN & 512 & 512 & 4 & 64 & \\
 &  & DGCNN & 512 & 1024 & 4 & 64 & \\
 &  & Linear & 2304 & 256 & - & - & \\
\hline 
\end{tabular}
\end{threeparttable}
\end{table}
\textbf{Patch Embedding} is a 1024 dimensional output of dVAE decoder (refer: dVAE architecture used Table \ref{tab:patch-encoder} and Subsection \ref{subsec:patch-point-cloud})
This dVAE model contains an encoder and a decoder trained to reconstruct patch data \cite{yu2022point} using Chamfer and KL Divergence loss.
The objective of the encoder is to create an 8192-dimensional token embedding for each patch, then the decoder scales down each token embedding to a 256-dimensional token embedding passing to the MLP to reconstruct the input patches. (For more details refer to Supplementary Material)
We extract Patch Embedding at 1024 dimensional linear layer, and concat with other embeddings at \textit{MECL}.

\subsection{Training Strategy} \label{subsec:training-tractoembed}
\textbf{TractoEmbed} concatenates all three embeddings—Streamline, Cluster, and Patch-- at the \textit{\textbf{MECL} (Multi-Embedding Concat Layer)} to feed the classifier MLP. During the training process, the Streamline and Patch Encoders, pre-trained on tasks mentioned above (refer to Section \ref{sec:method}), are kept non-trainable, while the Cluster Encoder and Classifier MLP are trainable. This is trained for 40 epochs at an initial learning rate of 0.0001 with a Cosine Annealing Warm Restarts learning rate scheduler using \textit{Focal Loss} to address class imbalance in major and minor tracts. Experimentally, we observed that concatenating the embeddings outperforms adding or merging them and focal loss performs better than cross entropy loss with these many classes. TractoEmbed extracts these multi-level embeddings to holistically represent streamlines and regional anatomy. 

\section{Results \& Discussions} \label{sec:results}
In this section, we present extensive ablation studies, and comparative results highlighting the effectiveness of our embeddings and data representations across different datasets. We evaluated all the results on the test split containing 20 subjects from a sample of 100 subjects. Classification Report with Accuracy and F1 scores for each class is described in the \textit{Supplementary Material}.

\begin{table}[!hbt]
\centering
\caption{\textbf{Comparative Results}: Model and Architecture performance across different Data Representations with a comparison of results with other state-of-the-art methods. Results for methods are sourced from \cite{xue2023tractcloud}, to eliminate discrepancies due to differences in training methods, except for \textit{ Hyperlocal PCD}. For TractCloud, hyperlocal PCD is made by setting $k_{local}$:5 and $k_{global}$: 0. All the experiments are done with keeping with their prescribed configurations static and only changing the input data.}
\label{tab:results}

\begin{tabular}{llcc}
\toprule
\textbf{Data}  & \textbf{Model: Type} & \textbf{Acc (\%)} & \textbf{F1 (\%)} \\
\midrule
\multirow{4}{*}{\textbf{Single Streamline}}  &  DeepWMA (CNN) & 90.29 & 88.12 \\
&  DCNN++ (CNN) & 91.26 & 89.14 \\
&  PointNet (PCD) & 91.36 & 89.12 \\
&  DGCNN (Graph) & \textbf{91.85} & \textbf{89.78} \\
\midrule
\multirow{3}{*}{\textbf{Local PCD} (k = 20)}  &  TractCloud: PointNet  & 91.51 & 89.25 \\
&  TractCloud: DGCNN (Graph) & 91.91 & 90.03 \\
&  \textbf{TractoEmbed} (ours) & \textbf{92.09} & \textbf{90.07} \\
\midrule
\multirow{2}{*}{\textbf{Hyperlocal PCD} (k = 5)}  &  TractCloud (PointNet) & 91.12 & 88.66 \\
&  \textbf{TractoEmbed} (ours) & \textcolor{red}{\textbf{93.04}} & \textcolor{red}{\textbf{91.38}} \\
\midrule
\multirow{2}{*}{\textbf{Local + Global Representation}}  &  TractCloud: PointNet & \textbf{92.28} & \textbf{90.36} \\
&  TractCloud: DGCNN (Graph) & 91.99 & 90.10 \\
\bottomrule
\end{tabular}
\end{table}

We present a comparison with several models, including DeepWMA \cite{zhang2020deep}, DCNN++ \cite{xu2019objective}, basic PointNet \cite{qi2017pointnet}, and DGCNN, using Single Streamline data. In Local PCD, TractoEmbed outperforms both variations of TractCloud \cite{xue2023tractcloud}.
In Hyperlocal PCD, we see that with only similar streamlines in the point cloud, TractoEmbed performs better than its performance in Local PCD, due to extensive focus on learning shape information of streamlines, as shown in Table \ref{tab:ablation-table}.

The comparative results in Table \ref{tab:results} highlight the efficacy and superior accuracy of our TractoEmbed framework for streamline classification and tract segmentation. Where TractCloud \cite{xue2023tractcloud} relies on local and global streamlines using a PointNet model to achieve registration-free tract segmentation TractoEmbed performs better even with spatially sparse hyperlocal PCD.
\begin{table}[!hbt]
    \centering
    \caption{\textbf{Ablation study} across a combination of embeddings with varying input point cloud densities to study their effect on Model Performance and finding the optimal hyperparameters (also refer Fig. \ref{fig:data-representation}). Here, $n_c$ number of points are randomly sampled from the total available points to make cluster data.}
    \label{tab:ablation-table}
    \begin{tabularx}{\textwidth}{p{3.5cm}|X|X|X|X|X|X}
        \hline
        \textbf{Multi Embeddings} & \textbf{Metric (\%)} & \multicolumn{3}{c|}{\textbf{Hyperlocal PCD (k=5)}} & \multicolumn{2}{c}{\textbf{Local PCD (k=20)}} \\
        \cline{3-7}
         & & \textbf{$n_c= $190 points} & \textbf{$n_c=$220 points} & \textbf{$n_c=$240 points} & \textbf{$n_c=$190 points} & \textbf{$n_c=$240 points} \\
        \hline
        \hline
        \textbf{cluster + streamline} & Acc & 92.917 & \textbf{93.038} & \textbf{93.020} & \textbf{91.494} &\textbf{91.383} \\
        & F1 & 91.198 & \textbf{91.381} & \textbf{91.346} & \textbf{89.338} & \textbf{89.239} \\
        \hline
        \textbf{cluster + patch} & Acc & 92.078 & 91.90 & 90.94 &81.502 & 80.451 \\
        & F1 & 89.891 & 89.574 & 88.489 & 74.765 &72.799 \\
        \hline
        \textbf{streamline + patch} & Acc & 91.654 & 91.065 & 89.675 & 91.165 & 90.956 \\
        & F1 & 89.525 & 88.97 & 87.331 &88.991 &88.781  \\
        \hline
        \textbf{cluster + patch } & Acc & \textbf{92.946} & 92.837 & 92.876 & \textbf{91.431 }&\textbf{91.409}  \\
        \textbf{+ streamline} & F1 & \textbf{91.284} & 91.091 & 91.164 & \textbf{89.239} & \textbf{89.36} \\
        \hline
    \end{tabularx}
\end{table}
The ablation study presented in Table \ref{tab:ablation-table} reveals the effectiveness of combining multiple embeddings used by TractoEmbed. As the neighboring point cloud becomes sparser, the representations need to be denser, increasing the need for more embeddings. Conversely, when the neighboring point cloud has a higher density of points, a pair of embeddings, cluster and streamline embedding, can achieve satisfactory performance. 
These findings ascertain the importance of incorporating dense streamline data representations from various perspectives/levels, including self, region, and neighbors.

\begin{table}[!htp]
\centering
\caption{\textbf{Ablation Study} showcasing the performance of individual encoders on different data representations for the tract segmentation through streamline classificati}
\label{tab:encoder-only}
\begin{tabular}{lllccc}
\toprule
\multirow{2}{*}{\textbf{Encoders only}} & \multirow{2}{*}{\textbf{TractoEmbed }} & \multirow{2}{*}{\textbf{ (\%)}} & \multicolumn{3}{c}{\textbf{Data}} \\\cmidrule{4-6}
& Models& & \textbf{Streamline} & \textbf{Cluster} & \textbf{Patch} \\
\midrule
\multirow{2}{*}{\textbf{Streamline Encoder}}&  \multirow{2}{*}{\textbf{CNN}} & Acc & 91.024& \textbf{-} & \textbf{-} \\
& & F1 & 88.894  & \textbf{-} & \textbf{-} \\
\hline
\multirow{2}{*}{\textbf{Cluster Encoder}}&  \multirow{2}{*}{\textbf{PointNet}} & Acc & 91.36 &90.54 & \textbf{-} \\
& & F1 & 89.12 & 88.31 & \textbf{-} \\
\hline
\multirow{2}{*}{\textbf{Patch Encoder}} &  \multirow{2}{*}{\textbf{dVAE}} & Acc & \textbf{-} & \textbf{-} & 85.87 \\
& & F1 & \textbf{-} & \textbf{-} & 82.48 \\
\bottomrule
\end{tabular}
\end{table}
The efficacy of a combination of embeddings can further be proven vital in increasing streamline classification accuracy as individual encoders perform poorly when compared to a combination of these embeddings (see Tables \ref{tab:encoder-only}  \ref{tab:ablation-table}).
Diving even further, there are slight improvements in F1 scores for projection fibers, striato-thalamo-pallido bundles, as observed in the Classification Report (refer to \textit{Supplementary Material}), indicating that Patch Embedding can effectively make information-dense patches of an input point cloud. 
Having multiple embeddings decreases the over-reliance on one knowledge representation, and makes TractoEmbed robust to changes in either of the representations. Explicit addition of Streamline Embedding containing information on the order of points and intra-streamline spatial information makes TractoEmbed robust to point cloud perturbations in Cluster Encoder. 

In summary, TractoEmbed demonstrates effectiveness in hyperlocal point clouds (regional examinations) and time-critical settings where a specific 3D brain segment is considered. It is also effective particularly for classifying structurally similar, minor, and projection fibers, achieving increased F1 scores and improved overall accuracy compared to LSTM-based approaches. TractoEmbed emphasizes the significance of fusing dense representations, incorporating various perspectives, including self, regional patches, and neighboring streamlines, which is crucial for extracting multiple types of information from low-fidelity streamline data. Future research can explore additional encoders, refined embedding combinations, optimal hyper-parameters, and different data representations. Also, there lies scope for improvement in finding unified models that can discriminate among highly similar streamlines or point clouds, with more classes.

\section{Conclusion}
With \textbf{TractoEmbed} we introduce an innovative method for Tract Segmentation, characterized by substantial accuracy, robustness, and modularity improvements. Our method integrates multiple embeddings from task-specific encoders to provide rich representations of streamlines, enabling a reduction in spatial input data requirements. It also demonstrates effectiveness in special cases, classifying structurally similar, minor, and projection fibers, by incorporating various data perspectives and minimal reliance on a single embedding. TractoEmbed also gives researchers the freedom to directly experiment with embeddings and data representations to get even better results.
With its spatially minimal data requirements, TractoEmbed can be useful for focused ROI-specific, and time-sensitive clinical settings. 
\textit{Code will be made available upon request.}

\section{Acknowledgement}
This research was supported by SERB Core Research Grant Project No:\\
CRG/2020/005492, IIT Mandi.

\bibliographystyle{splncs04}
\bibliography{references}

\begin{thebibliography}{10}
\providecommand{\url}[1]{\texttt{#1}}
\providecommand{\urlprefix}{URL }
\providecommand{\doi}[1]{https://doi.org/#1}

\bibitem{basser1994mr}
Basser, P.J., Mattiello, J., LeBihan, D.: Mr diffusion tensor spectroscopy and imaging. Biophysical journal  \textbf{66}(1),  259--267 (1994)

\bibitem{basser2000vivo}
Basser, P.J., et~al.: In vivo fiber tractography using dt-mri data. Magnetic resonance in medicine  \textbf{44}(4),  625--632 (2000)

\bibitem{behrens2007probabilistic}
Behrens, T.E., et~al.: Probabilistic diffusion tractography with multiple fibre orientations: What can we gain? neuroimage  \textbf{34}(1),  144--155 (2007)

\bibitem{berto2021classifyber}
Bert{\`o}, G., et~al.: Classifyber, a robust streamline-based linear classifier for white matter bundle segmentation. NeuroImage  \textbf{224},  117402 (2021)

\bibitem{dumais2023fiesta}
Dumais, F., et~al.: Fiesta: Autoencoders for accurate fiber segmentation in tractography. NeuroImage  \textbf{279},  120288 (2023)

\bibitem{edwards2022developing}
Edwards, A.D., et~al.: The developing human connectome project neonatal data release. Frontiers in neuroscience  \textbf{16},  886772 (2022)

\bibitem{funk2023humans}
Funk, A.T., Hassan, A.A., Br{\"u}ggemann, N., Sharma, N., Breiter, H.C., Blood, A.J., Waugh, J.L.: In humans, striato-pallido-thalamic projections are largely segregated by their origin in either the striosome-like or matrix-like compartments. Frontiers in neuroscience  \textbf{17},  1178473 (2023)

\bibitem{garyfallidis2018recognition}
Garyfallidis, E., C{\^o}t{\'e}, M.A., Rheault, F., Sidhu, J., Hau, J., Petit, L., Fortin, D., Cunanne, S., Descoteaux, M.: Recognition of white matter bundles using local and global streamline-based registration and clustering. NeuroImage  \textbf{170},  283--295 (2018)

\bibitem{garyfallidis2012quickbundles}
Garyfallidis, E., et~al.: Quickbundles, a method for tractography simplification. Frontiers in neuroscience  \textbf{6}, ~175 (2012)

\bibitem{gupta2017brainsegnet}
Gupta, T., Patil, S.M., Tailor, M., Thapar, D., Nigam, A.: Brainsegnet: a segmentation network for human brain fiber tractography data into anatomically meaningful clusters. arXiv preprint arXiv:1710.05158  (2017)

\bibitem{jha2019fs2net}
Jha, R.R., Patil, S., Nigam, A., Bhavsar, A.: Fs2net: fiber structural similarity network (fs2net) for rotation invariant brain tractography segmentation using stacked lstm based siamese network. In: Computer Analysis of Images and Patterns: 18th International Conference, CAIP 2019, Salerno, Italy, September 3--5, 2019, Proceedings, Part II 18. pp. 459--469. Springer (2019)

\bibitem{lam2018trafic}
Lam, P.D.N., Belhomme, G., Ferrall, J., Patterson, B., Styner, M., Prieto, J.C.: Trafic: fiber tract classification using deep learning. In: Medical Imaging 2018: Image Processing. vol. 10574, pp. 257--265. SPIE (2018)

\bibitem{liu2019deepbundle}
Liu, F., Feng, J., Chen, G., Wu, Y., Hong, Y., Yap, P.T., Shen, D.: Deepbundle: fiber bundle parcellation with graph convolution neural networks. In: Graph Learning in Medical Imaging: First International Workshop, GLMI 2019, Held in Conjunction with MICCAI 2019, Shenzhen, China, October 17, 2019, Proceedings 1. pp. 88--95. Springer (2019)

\bibitem{lucena2022informative}
Lucena, O., et~al.: Informative and reliable tract segmentation for preoperative planning. Frontiers in radiology  \textbf{2},  866974 (2022)

\bibitem{malcolm2009neural}
Malcolm, J.G., Shenton, M.E., Rathi, Y.: Neural tractography using an unscented kalman filter. In: International Conference on Information Processing in Medical Imaging. pp. 126--138. Springer (2009)

\bibitem{marek2011parkinson}
Marek, K., et~al.: The parkinson progression marker initiative (ppmi). Progress in neurobiology  \textbf{95}(4),  629--635 (2011)

\bibitem{o2007automatic}
O'Donnell, L.J., Westin, C.F.: Automatic tractography segmentation using a high-dimensional white matter atlas. IEEE transactions on medical imaging  \textbf{26}(11),  1562--1575 (2007)

\bibitem{o2012unbiased}
O’Donnell, L.J., Wells, W.M., Golby, A.J., Westin, C.F.: Unbiased groupwise registration of white matter tractography. In: Medical Image Computing and Computer-Assisted Intervention--MICCAI 2012: 15th International Conference, Nice, France, October 1-5, 2012, Proceedings, Part III 15. pp. 123--130. Springer (2012)

\bibitem{qi2017pointnet}
Qi, Charles, o.: Pointnet: Deep learning on point sets for 3d classification and segmentation. In: Proceedings of the IEEE conference on computer vision and pattern recognition. pp. 652--660 (2017)

\bibitem{rolfe2016discrete}
Rolfe, J.T.: Discrete variational autoencoders. arXiv preprint arXiv:1609.02200  (2016)

\bibitem{st2022fast}
St-Onge, E., et~al.: Fast streamline search: an exact technique for diffusion mri tractography. Neuroinformatics  \textbf{20}(4),  1093--1104 (2022)

\bibitem{tournier2007robust}
Tournier, J.D., Calamante, F., Connelly, A.: Robust determination of the fibre orientation distribution in diffusion mri: non-negativity constrained super-resolved spherical deconvolution. Neuroimage  \textbf{35}(4),  1459--1472 (2007)

\bibitem{tournier2004direct}
Tournier, J.D., et~al.: Direct estimation of the fiber orientation density function from diffusion-weighted mri data using spherical deconvolution. Neuroimage  \textbf{23}(3),  1176--1185 (2004)

\bibitem{van2013wu}
Van~Essen, D.C., et~al.: The wu-minn human connectome project: an overview. Neuroimage  \textbf{80},  62--79 (2013)

\bibitem{vindas2023geolab}
Vindas, N., et~al.: Geolab: Geometry-based tractography parcellation of superficial white matter. In: 2023 IEEE 20th International Symposium on Biomedical Imaging (ISBI). pp.~1--5. IEEE (2023)

\bibitem{volkow2018conception}
Volkow, N.D., et~al.: The conception of the abcd study: From substance use to a broad nih collaboration. Developmental cognitive neuroscience  \textbf{32}, ~4--7 (2018)

\bibitem{wang2022accurate}
Wang, Z., et~al.: Accurate corresponding fiber tract segmentation via fibergeomap learner. In: International Conference on Medical Image Computing and Computer-Assisted Intervention. pp. 143--152. Springer (2022)

\bibitem{wasserthal2018tractseg}
Wasserthal, J., Neher, P., Maier-Hein, K.H.: Tractseg-fast and accurate white matter tract segmentation. NeuroImage  \textbf{183},  239--253 (2018)

\bibitem{xu2019objective}
Xu, H., Dong, M., Lee, M.H., O’Hara, N., Asano, E., Jeong, J.W.: Objective detection of eloquent axonal pathways to minimize postoperative deficits in pediatric epilepsy surgery using diffusion tractography and convolutional neural networks. IEEE transactions on medical imaging  \textbf{38}(8),  1910--1922 (2019)

\bibitem{xue2023tractcloud}
Xue, T., et~al.: Tractcloud: Registration-free tractography parcellation with a novel local-global streamline point cloud representation. In: International Conference on Medical Image Computing and Computer-Assisted Intervention. pp. 409--419. Springer (2023)

\bibitem{yu2022point}
Yu, X., Tang, L., Rao, Y., Huang, T., Zhou, J., Lu, J.: Point-bert: Pre-training 3d point cloud transformers with masked point modeling. In: Proceedings of the IEEE/CVF conference on computer vision and pattern recognition. pp. 19313--19322 (2022)

\bibitem{zhang2018anatomically}
Zhang, F., Wu, Y., Norton, I., Rigolo, L., Rathi, Y., Makris, N., O'Donnell, L.J.: An anatomically curated fiber clustering white matter atlas for consistent white matter tract parcellation across the lifespan. Neuroimage  \textbf{179},  429--447 (2018)

\bibitem{zhang2020deep}
Zhang, F., et~al.: Deep white matter analysis (deepwma): Fast and consistent tractography segmentation. Medical Image Analysis  \textbf{65},  101761 (2020)

\end{thebibliography}

\end{document}


\titlerunning{TractoEmbed}
\authorrunning{Anoushkrit Goel et al.}
\authorrunning{A. Goel et al.}
\label{suppl}
\begin{center}
      \vspace*{-22pt}
        \begin{tabular}[t]{c}{
        \large\bf
              Supplementary Material
            }
        \end{tabular}
        \par
\end{center}
\textbf{RAS Coordinates}
The RAS (Right-Anterior-Superior) coordinate system is a standard convention in medical imaging, where the x-axis increases from left to right, the y-axis increases from posterior to anterior, and the z-axis increases from inferior to superior.

\section{Patch Encoder Training} 
The discrete Variational Autoencoder (dVAE) model consists of an encoder and a decoder, these models are trained to reconstruct patch data using Chamfer Distance and KL Divergence loss. Initially, the patch data sampled using FPS (Farthest Point Sampling) and kNN (k-Nearest Neighbours) is of shape (64, 16, 3) where dimensions represent (number of patches, number of points per patch, number of features). Here, (64,16,3) vector is fed to a mini PointNet Model, resulting in a (64, 256)-dimensional embedding. 

This embedding goes through Conv1D and DGCNN encoder, yielding a (64, 8192)-dimensional latent space where each patch is projected to 8192 dimensions. To introduce stochasticity, this 8192-dimensional latent space undergoes Gumbel Softmax for categorical sampling and then MLP, reshaping it to (64, 256). 
Output of encoder passed to the DGCNN decoder, an MLP and a Convolutional decoder to reconstruct the original point cloud (as described in Table 4, also refer \textit{Yu et al. 2022})

Training involves a combination loss function that fuses reconstruction loss on Chamfer Distance L1 and KL divergence on a uniform distribution. 
The loss function is defined as 
$$loss= loss_1 + kld_{weight} * loss_2$$
(where $loss_1$: ChamferDistL1;  $loss_2$:KL divergence). 

Initially, $kld_{weight}$ is set to 0 for the first 10,000 iterations, and then increases with a cosine curve reaching 0.1 over the next 100,000 iterations.
$$kld_{weight} = 0.1 - 0.1 * (1 + cos(pi * (n_{itr} - 10000) / 100000))$$

The optimization uses the AdamW optimizer with an initial learning rate of 0.0005 and weight decay of 0.0005. 
The learning rate is managed by a Cosine Learning Rate Scheduler with parameters: $t_{initial}$ = 300, $t_{mul}$ = 1, $lr_{min}$ = 1e-6, $decay_{rate}$ = 0.1, $warmupLR_{init}$ = 1e-6, $warmup_t$ = 10, $cycle_{limit}$ = 1, $t_{epochs}$ = True.
\newpage

\section{Classification Report}
\begin{longtable}{p{6cm}|c|c|c|c}
    
    \caption{\textbf{Classification Report} on Test Data of 20 subjects spanning from different datasets} 
    \label{tab:classification-report} \\
    \hline
    \textbf{Tract} & \textbf{Precision} & \textbf{Recall} & \textbf{f1-score} & \textbf{Support} \\
    \hline
    \endfirsthead

    \hline
    \textbf{Tract} & \textbf{Precision} & \textbf{Recall} & \textbf{f1-score} & \textbf{Support} \\
    \hline
    \endhead

    \hline
    \endfoot
    \hline
    \endlastfoot
    (\textbf{AF}) arcuate fasciculus & 0.91125 & 0.95624 & 0.93320 & 2491 \\
    (\textbf{CB}) cingulum bundle & 0.90006 & 0.95009 & 0.92440 & 3346 \\
    (\textbf{EC}) external capsule & 0.81636 & 0.88314 & 0.84844 & 599 \\
    (\textbf{EmC}) extreme capsule & 0.85696 & 0.91240 & 0.88381 & 742 \\
    (\textbf{ILF}) inferior longitudinal fasciculus & 0.90465 & 0.93620 & 0.92015 & 3699 \\
    (\textbf{IOFF}) inferior occipito-frontal fasciculus & 0.94879 & 0.95342 & 0.95110 & 2662 \\
    (\textbf{MdLF}) middle longitudinal fasciculus & 0.91158 & 0.93352 & 0.92242 & 3490 \\
    (\textbf{SLF-I}) superior longitudinal fasciculus I & 0.90785 & 0.93904 & 0.92318 & 2969 \\
    (\textbf{SLF-II}) superior longitudinal fasciculus II & 0.91819 & 0.93073 & 0.92442 & 3075 \\
    (\textbf{SLF-III}) superior longitudinal fasciculus III & 0.89665 & 0.92775 & 0.91193 & 1384 \\
    (\textbf{UF}) uncinate fasciculus & 0.88528 & 0.94748 & 0.91532 & 1409 \\
    (\textbf{CST}) corticospinal tract & 0.90917 & 0.95804 & 0.93297 & 2121 \\
    (\textbf{CR-F}) corona radiata-frontal & 0.91929 & 0.93720 & 0.92816 & 2054 \\
    (\textbf{CR-P}) corona radiata-parietal & 0.88684 & 0.93204 & 0.90888 & 412 \\
    (\textbf{SF}) striato-frontal & 0.86588 & 0.89665 & 0.88100 & 3348 \\
    (\textbf{SO}) striato-occipital & 0.85714 & 0.86667 & 0.86188 & 270 \\
    (\textbf{SP}) striato-parietal & 0.84211 & 0.83436 & 0.83821 & 326 \\
    (\textbf{TF}) thalamo-frontal & 0.92455 & 0.94921 & 0.93672 & 4725 \\
    (\textbf{TO}) thalamo-occipital & 0.89973 & 0.91586 & 0.90772 & 725 \\
    (\textbf{TT}) thalamo-temporal & 0.89396 & 0.92885 & 0.91107 & 2024 \\
    (\textbf{TP}) thalamo-parietal & 0.84007 & 0.88267 & 0.86084 & 1696 \\
    (\textbf{PLIC}) posterior limb of internal capsule & 0.88182 & 0.88584 & 0.88383 & 438 \\
    (\textbf{CC1}) corpus callosum 1 & 0.82748 & 0.85197 & 0.83955 & 304 \\
    (\textbf{CC2}) corpus callosum 2 & 0.93202 & 0.95149 & 0.94165 & 2824 \\
    (\textbf{CC3}) corpus callosum 3 & 0.93510 & 0.94764 & 0.94133 & 1566 \\
    (\textbf{CC4}) corpus callosum 4 & 0.94894 & 0.96086 & 0.95486 & 1354 \\
    (\textbf{CC5}) corpus callosum 5 & 0.92639 & 0.94681 & 0.93649 & 1316 \\
    (\textbf{CC6}) corpus callosum 6 & 0.91497 & 0.96450 & 0.93908 & 2789 \\
    (\textbf{CC7}) corpus callosum 7 & 0.93963 & 0.90967 & 0.92441 & 941 \\
    (\textbf{CPC}) cortico-ponto-cerebellar & 0.86104 & 0.91557 & 0.88747 & 379 \\
    (\textbf{ICP}) inferior cerebellar peduncle & 0.92437 & 0.94624 & 0.93518 & 465 \\
    Intra-CBLM-I-P & 0.88861 & 0.92423 & 0.90607 & 2270 \\
    Intra-CBLM-PaT & 0.94334 & 0.97983 & 0.96124 & 5998 \\
    (\textbf{MCP}) middle cerebellar peduncle & 0.93772 & 0.94784 & 0.94275 & 1112 \\
    (\textbf{Sup-F}) & 0.92529 & 0.95181 & 0.93836 & 16187 \\
    (\textbf{Sup-FP}) & 0.88692 & 0.91733 & 0.90187 & 2129 \\
    (\textbf{Sup-O}) & 0.90185 & 0.93723 & 0.91920 & 1147 \\
    (\textbf{Sup-OT}) & 0.88313 & 0.91682 & 0.89966 & 1599 \\
    (\textbf{Sup-P}) & 0.92276 & 0.93842 & 0.93053 & 7486 \\
    (\textbf{Sup-PO}) & 0.90145 & 0.92703 & 0.91406 & 2220 \\
    (\textbf{Sup-PT}) & 0.91627 & 0.94441 & 0.93012 & 5666 \\
    (\textbf{Sup-T}) & 0.88718 & 0.92335 & 0.90491 & 3066 \\
    Other & 0.95230 & 0.91948 & 0.93560 & 95177 \\
\end{longtable}
\bibliography{references}